\documentclass[letterpaper, 10 pt, twoside, conference]{ieeeconf}  % 
\IEEEoverridecommandlockouts                              % 
\usepackage{amsmath} % assumes amsmath package installed
\usepackage{amssymb}  % assumes amsmath package installed
\usepackage[english]{babel}
\usepackage[utf8]{inputenc}
\usepackage{graphicx}
\usepackage{cite}
\title{\LARGE \bf
SPECFACE - A Dataset of Human Faces Wearing Spectacles 
}

\author{Anirban Dasgupta, Shubhobrata Bhattacharya and Aurobinda Routray\\Indian Institute of Technology Kharagpur\\India}% <-this % stops a space
\begin{document}
\pagestyle{headings}
\maketitle
\markboth{Preprint Submitted to Arxiv}{To appear in IEEE Students' Technology Symposium 2016}
\thispagestyle{empty}
\pagestyle{empty}
\begin{abstract}
This paper presents a database of human faces for persons wearing spectacles. The database consists of images of faces having significant variations with respect to illumination, head pose, skin color, facial expressions and sizes, and nature of spectacles. The database contains data of 60 subjects. This database is expected to be a precious resource for the development and evaluation of algorithms for face detection, eye detection, head tracking, eye gaze tracking, etc., for subjects wearing spectacles. As such, this can be a valuable contribution to the computer vision community.
\end{abstract}
\section{Introduction}
\begin{figure}[h]
\centering
\includegraphics[width=0.6\linewidth]{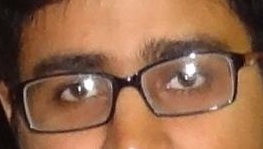}
\caption{Glint on eye due to spectacles}
\label{glint}
\end{figure}

In the field of computer vision, face detection \cite{gupta2011analysis}, face recognition, facial expression classification \cite{happy2013automated}, eye detection \cite{sengupta2016alertness}, head tracking, eye gaze tracking \cite{hjelmaas2001face} etc. has gained much popularity . One major bottleneck in such algorithms is that they show limited accuracies when the face of the target user is occluded by spectacles \cite{dasgupta2013board}. Several researches have been developing algorithms for ocular feature computation and identification, which address the issue of transparent spectacles. Singvi \textit{et al.} \cite{singvi2012real} have developed an algorithm for the detection of presence of spectacles in a face. The scope of their work is limited to the detection of spectacles. They do not address the issue of eye information classification on such images, as they conclude that spectacles deform the information content of the eye due to glint. For example, let us consider the case of classifying the eye as open or closed. If the person is wearing spectacles, the glass will acts as a reflector, thereby the information content related to the state of eye closure is difficult to estimate using the state-of-the art algorithms. As an example consider the situation in Fig. \ref{glint}, where we can see how a glare can destroy information and introduce spurious features.\\Orazio \textit{et al.} \cite{orazio2004algorithm} have proposed a real-time algorithm for eye detection using geometrical information of the iris. Their algorithm shows limited accuracy with people wearing glasses as the image of the iris region may be deformed due to glint. Asteriadis \textit{et al.} \cite{asteriadis2006eye} have used pixel to edge information for localization of the eye region in face images. As evident, presence of glasses cause a hindrance in obtaining a significantly accurate method, as glints may introduce spurious edges. Park \textit{et al.} \cite{park2005glasses} have devised a method for compensating the effect of spectacles. However, such algorithms which work with occlusion with spectacles do not have a standard platform to compare due to the unavailability of exclusively dedicated image databases of users wearing spectacles.\\In this work, we try to facilitate such algorithms, by creating a standard face database of human subjects wearing spectacles. We name this database SPECFACE (Spectacle Bearing Faces). The database consists of facial images prepared following the appropriate criteria as that of the Gold Standard Face databases such as the CMU database \cite{sim2002cmu} and the FERET database \cite{phillips1998feret}. The database not only can be used for eye information classification for users wearing spectacles, but also can be used for other applications such as face recognition, head pose estimation, eye detection, facial landmark detection, facial expression classification, gender classification etc.\\This paper is organized as follows. Some popular existing databases are discussed in Section II. Section III describes about the SPECFACE database content. Section IV treats a quality assessment of the database. Finally, Section V concludes the paper.

\section{Existing Databases}
There are some existing popular face databases which contain few images of spectacles. The Face Recognition Technology (FERET) database \cite{phillips1998feret} is a popular database which contain images of 1199 individuals. It has been a gold standard benchmark for face recognition. The CMU Pose, Illumination, and Expression Database \cite{sim2002cmu} comes in handy with 68 persons.  The Japanese female facial expression (JAFFE) database is a special database of female candidates where difference facial expressions pertaining to different emotions have been used. The JAFFE database contains 213 images of seven facial expressions posed by 10 Japanese female models. Recently, Happy \textit{et al.} have created a video database with near infrared (NIR) lighting in \cite{happy2012video}. The database consists of 60 subjects. The authors have employed Ga-As NIR LED's for illuminating the face. The major variations considered in the work were variations in NIR illumination, facial expressions, occlusion with spectacles as well as hands, head rotations (both off-plane and in-plane), etc. These databases contain some images of spectacles, without having glary noise. This is due to the acquisition in a controlled environment. There are no exclusive database available for spectacles with added glare noise, which make it difficult to benchmark eye detection algorithms for subjects wearing spectacles. This paper attempts the same with naturally occuring glare noise. The images are captured both at controlled as well as wild environments to mimic naturally occurring situations. The database can not only aid eye detection with glasses, but also can be used to benchmark algorithms on face detection, face recognition, gender classification, ethnicity classification, head pose estimation, facial expression recognition and many more.
\begin{figure}[h]
\includegraphics[width=\linewidth]{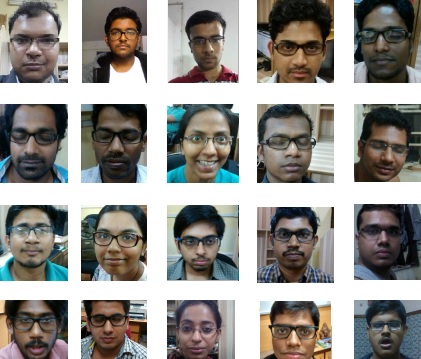}
\caption{Sample Images from SPECFACE}
\end{figure}
\section{Database Content}
This section discusses the protocol followed to created the database as well as its content. The variations along with their applications in concerned algorithms are tabulated in Table \ref{summary}.
\subsection{Imaging Device}
The images were recorded using a smart-phone camera using various resolutions. There were two different resolutions prominently used \textit{viz.} 1920$\times$1260 and 640$\times$480. The sensor is a CMOS sensor. The images are stored in JPEG compression. The distance from the camera varied from 50cm to 250cm.
\subsection{Age Group}
The database consists of people within the age group of 20 years to 60 years. This variation ensures that the dataset can facilitate algorithms which can work on persons with a varied age-group.
\subsection{Head Pose}
\subsubsection{Off-plane rotation}
\begin{figure}[h]
\includegraphics[width=\linewidth]{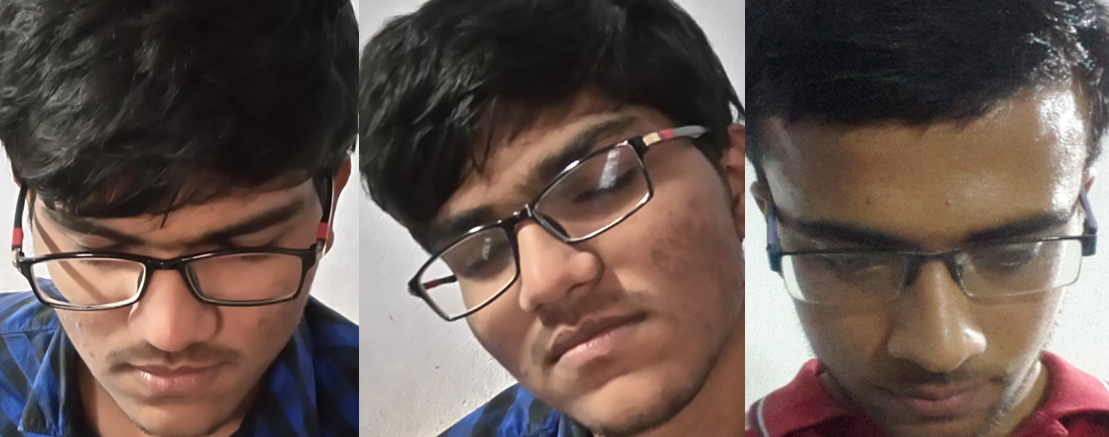}
\caption{Variations in head pose}
\label{pose}
\end{figure}

The participants were instructed to rotate  their faces towards right, left, up and down successively.  Fig. \ref{pose} shows some off-plane rotations. 
\subsubsection{In-plane rotation}
Subjects were instructed to tilt their head gradually in both
right-left directions to obtain in-plane rotated faces.
\subsubsection{Head nodding}
Head nodding in vertical, horizontal and random direction are also recorded to add some more variation in head
tracking. This causes motion blur which help is evaluating image stabilization algorithms.
\subsection{Illumination Variation}
The illumination levels used for preparing the dataset varied between 80 lux to 8000 lux. This variation was made as a natural setting to evaluate illumination-variation algorithms.
\begin{figure}[h]
\includegraphics[width=\linewidth]{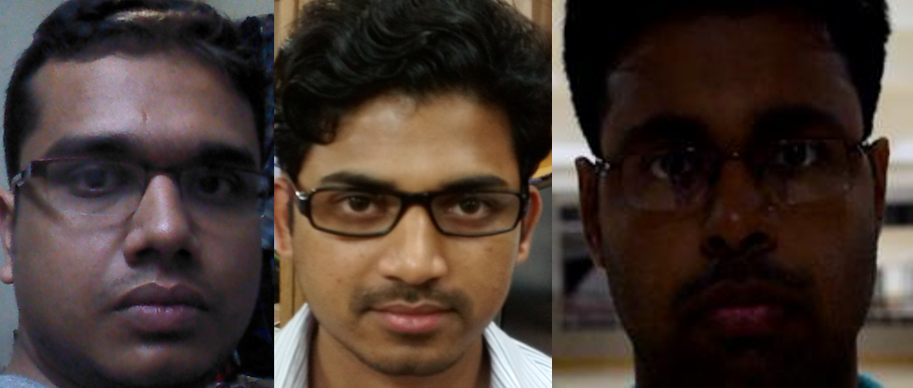}
\caption{Variations in Illumination}
\label{illumination}
\end{figure}
\subsection{Skin tone and hue}
\begin{figure}[h]
\includegraphics[width=\linewidth]{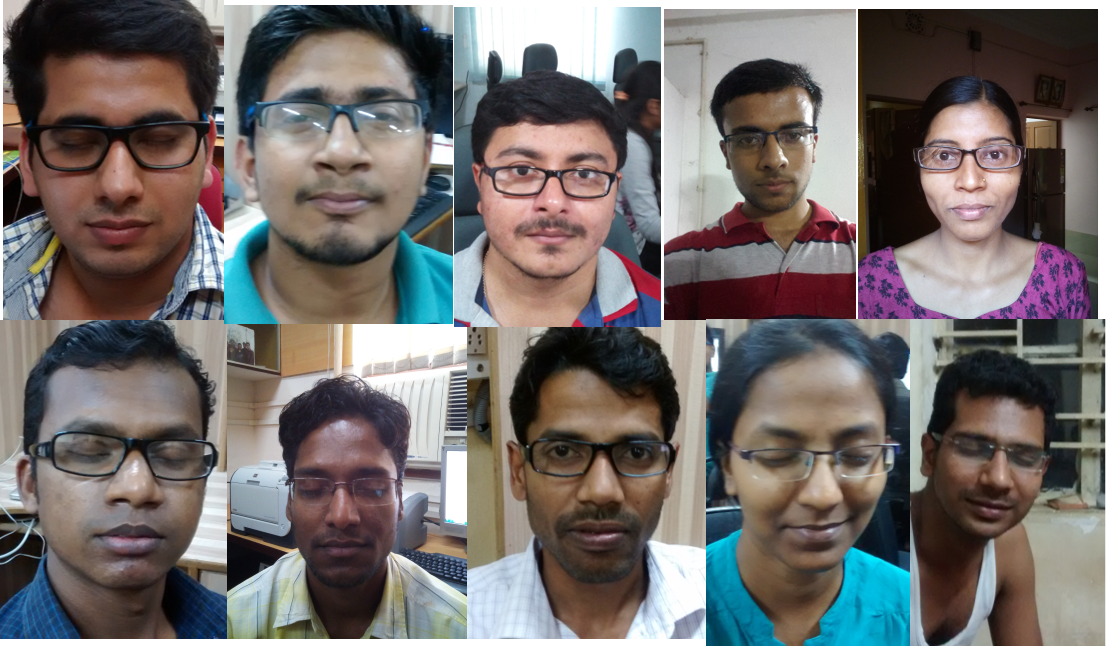}
\caption{Variations in Hue and Texture of Skin}
\label{hue}
\end{figure}

This variation has been made to consider the algorithms which uses skin texture and hue components of facial landmarks for face recognition. Fig. \ref{hue} shows some images of subjects having different skin tones.
\begin{figure*}
\includegraphics[width=\linewidth]{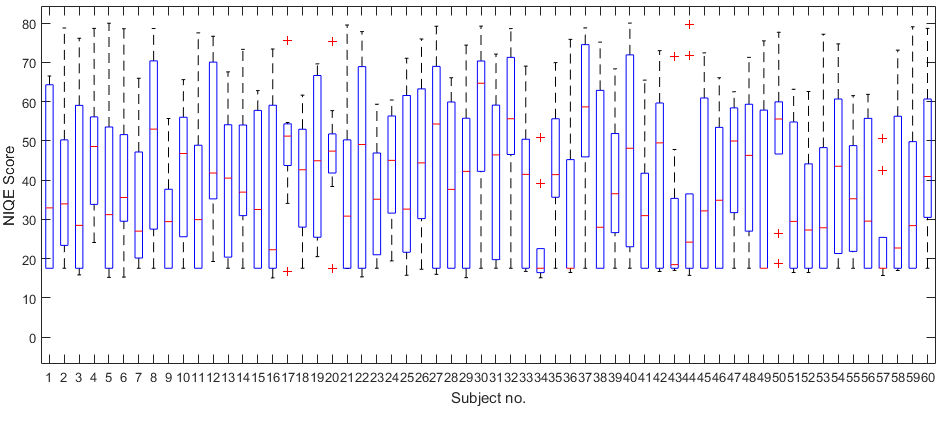}
\caption{Variations in NIQE scores}
\label{niqe}
\end{figure*}
\subsection{Gender Variation}
\begin{figure}[h]
\includegraphics[width=\linewidth]{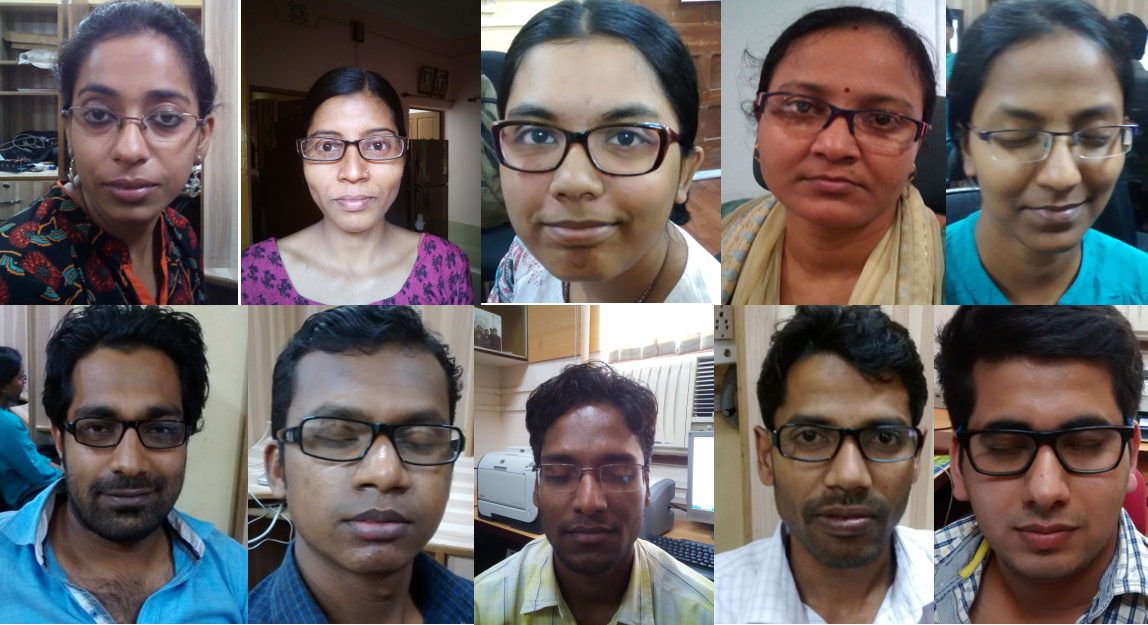}
\caption{Upper Female, lower male subjects}
\label{gender}
\end{figure}

This variation is considered for algorithms which classifies the gender of a person based on face images using facial features. Fig. \ref{gender} shows some images. The upper row contain female faces while the lower row contain male faces.
\subsection{Eye Closure States}
\begin{figure}[h]
\includegraphics[width=\linewidth]{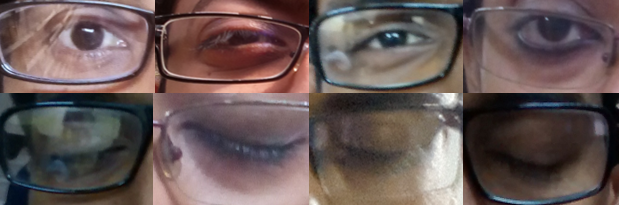}
\caption{Eye Closure States}
\label{eye}
\end{figure}

The SPECFACE dataset also contains open as well as closed eye, which helps in evaluating eye state classification algorithms. Such algorithms may be useful in computing blink rates or the assessment of drowsiness of a person \cite{dasgupta2013vision}. Fig. \ref{eye} shows some sample images of eyes open (top row) and closed (bottom row).

\section{Variation Assessment}
The variation assessment is required to assess the statistical importance of the dataset.
\subsection{Naturalness Image Quality Evaluator (NIQE)}
Natural Image Quality Evaluator (NIQE) blind image quality assessment is a blind image quality analyzer that only makes use of measurable deviations from statistical regularities observed in images, without any prior training on human-rated images \cite{galbally2014image}. It is based on a simple space domain natural scene statistic model. The results for NIQE on the dataset are computed and are shown in Fig. \ref{niqe} as a boxplot.
\begin{table}[]
\centering
\caption{Summary of variations considered for preparing SPECFACE with justifications}
\label{summary}
\begin{tabular}{|l|l|}
\hline
\textbf{Variation} & \textbf{Evaluation Algorithms}                                                                                    \\ \hline
Off-plane rotation & \begin{tabular}[c]{@{}l@{}}non-frontal face detection\\ and recognition\end{tabular}                              \\ \hline
In-plane rotation  & tilted face detection                                                                                             \\ \hline
Head Nodding       & image stabilization                                                                                               \\ \hline
Illumination       & illumination invariance                                                                                           \\ \hline
Age                & age estimation from face image                                                                                    \\ \hline
Gender             & \begin{tabular}[c]{@{}l@{}}facial features for gender\\ classification\end{tabular}                               \\ \hline
Skin tone and hue  & \begin{tabular}[c]{@{}l@{}}skin based face recognition,\\ ethnicity classification\end{tabular}                   \\ \hline
Eye closure        & \begin{tabular}[c]{@{}l@{}}eye state analysis, blink\\ frequency computation,\\ drowsiness detection\end{tabular} \\ \hline
\end{tabular}
\end{table}
\begin{figure*}
\includegraphics[width=\linewidth]{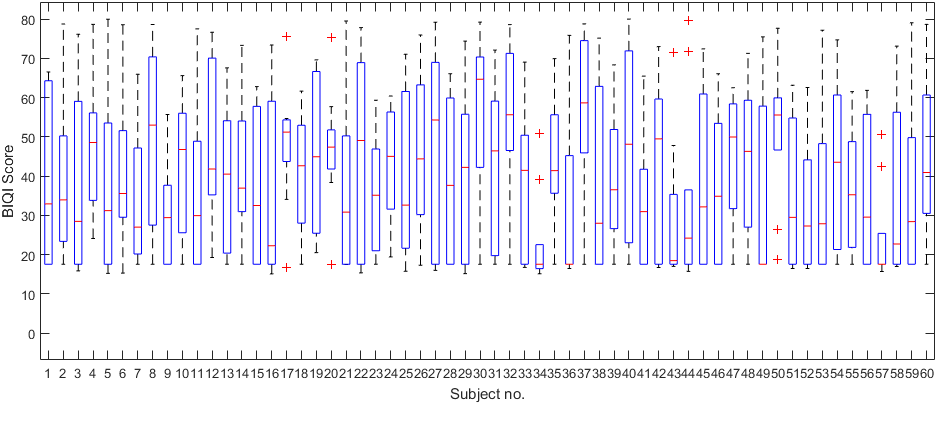}
\caption{Variations in BIQI scores}
\label{biqi}
\end{figure*}
\subsection{Blind Image Quality Index (BIQI)}
Blind Image Quality Index is a natural scene statistic -based distortion-generic blind image quality assessment model \cite{mittal2011blind}. It is based on predicting image quality based on observing
the statistics of local discrete cosine transform coefficients, and it requires only minimal training. The method is shown to correlate highly with human perception of quality.
The results are computed and are shown in Fig. \ref{biqi} as a boxplot. From the plots of NIQE and BIQI, it is evident that the dataset has significant variations in terms of image quality. Smaller values of BIQI and NIQE indicate that there are deliberately introduced poor quality images, which will be real test for highly advanced machine learning algorithms.
\section{Conclusion}
In this paper, we have contributed an image database to the computer vision community. This database aims to aid research on facial landmark detection, face recognition and allied areas, where spectacles form a limitation to existing algorithms. The database consists of images of 60 persons within an age-group of 20 to 60 years. The variations and image quality of the database are analysed using the BIQI and NIQE metrics. The database is expected to be a valuable asset to the computer vision community.

\subsection{Availability}
The database is available at the link $https://drive.google.com/drive/folders/0B75ThlVzcSP5W\\UxaOFVhQkVOVVE$. The link will be accessible to the concerned authority upon a request. An online form will be asked to fill, upon which the access will be granted. The database is expected to be 
\bibliographystyle{IEEEtran}
\bibliography{library}
\end{document}